\documentclass[sigconf]{acmart}

\AtBeginDocument{%
  \providecommand\BibTeX{{%
    \normalfont B\kern-0.5em{\scshape i\kern-0.25em b}\kern-0.8em\TeX}}}

\usepackage[symbol]{footmisc}

\usepackage{epsfig}
\usepackage{graphicx}
\usepackage{amsmath}
\usepackage{amssymb}

\usepackage{caption} 
\usepackage{subcaption}
\usepackage{graphicx}

\AtBeginDocument{%
  \providecommand\BibTeX{{%
    \normalfont B\kern-0.5em{\scshape i\kern-0.25em b}\kern-0.8em\TeX}}}

\renewcommand\footnotetextcopyrightpermission[1]{} 
\begin{document}


\fancyhead{}

\title{ Running Event Visualization \\ using Videos  from Multiple Cameras}

\author{Y. Napolean*}
  \thanks{*These authors contributed equally}
\affiliation{%
  \institution{TU Delft}
}
\email{y.napolean@tudelft.nl}

\author{P.T. Wibowo*}
\affiliation{%
  \institution{TU Delft}
}
\email{priaditeguhwibowo@student.tudelft.nl}

\author{J.C. van Gemert}
\affiliation{%
  \institution{TU Delft}
}
\email{j.c.vangemert@tudelft.nl}

\renewcommand{\shortauthors}{Wibowo, et al.}

\begin{abstract}
  Visualizing the trajectory of multiple runners with videos collected at different points in a race could be useful for sports performance analysis. The videos and the trajectories can also aid in athlete health monitoring. While the runners unique ID and their appearance are distinct, the task is not straightforward because the video data does not contain explicit information as to which runners appear in each of the videos. There is no direct supervision of the model in tracking athletes, only filtering steps to remove irrelevant detections. Other factors of concern include occlusion of runners and harsh illumination.  To this end, we identify two methods for runner identification at different points of the event, for determining their trajectory. One is scene text detection which recognizes the runners by detecting a unique 'bib number' attached to their clothes and the other is person re-identification which detects the runners based on their appearance. We train our method without ground truth but to evaluate the proposed methods, we create a ground truth database  which consists of video and frame interval information where the runners appear. The videos in the dataset was recorded by nine cameras at different locations during the a marathon event. This data is annotated with bib numbers of runners appearing in each video. The bib numbers of runners known to occur in the frame are used to filter irrelevant text and numbers detected. Except for this filtering step, no supervisory signal is used. The experimental evidence shows that the scene text recognition method achieves an F1-score of 74. Combining the two methods, that is - using samples collected by text spotter to train the re-identification model yields a higher F1-score of 85.8. Re-training the person re-identification model with identified inliers yields a slight improvement in performance(F1 score of 87.8). This combination of text recognition and person re-identification can be used in conjunction with video metadata to visualize running events.
\end{abstract}






\maketitle

\begin{figure}
  \centering
  \includegraphics[scale=0.3]{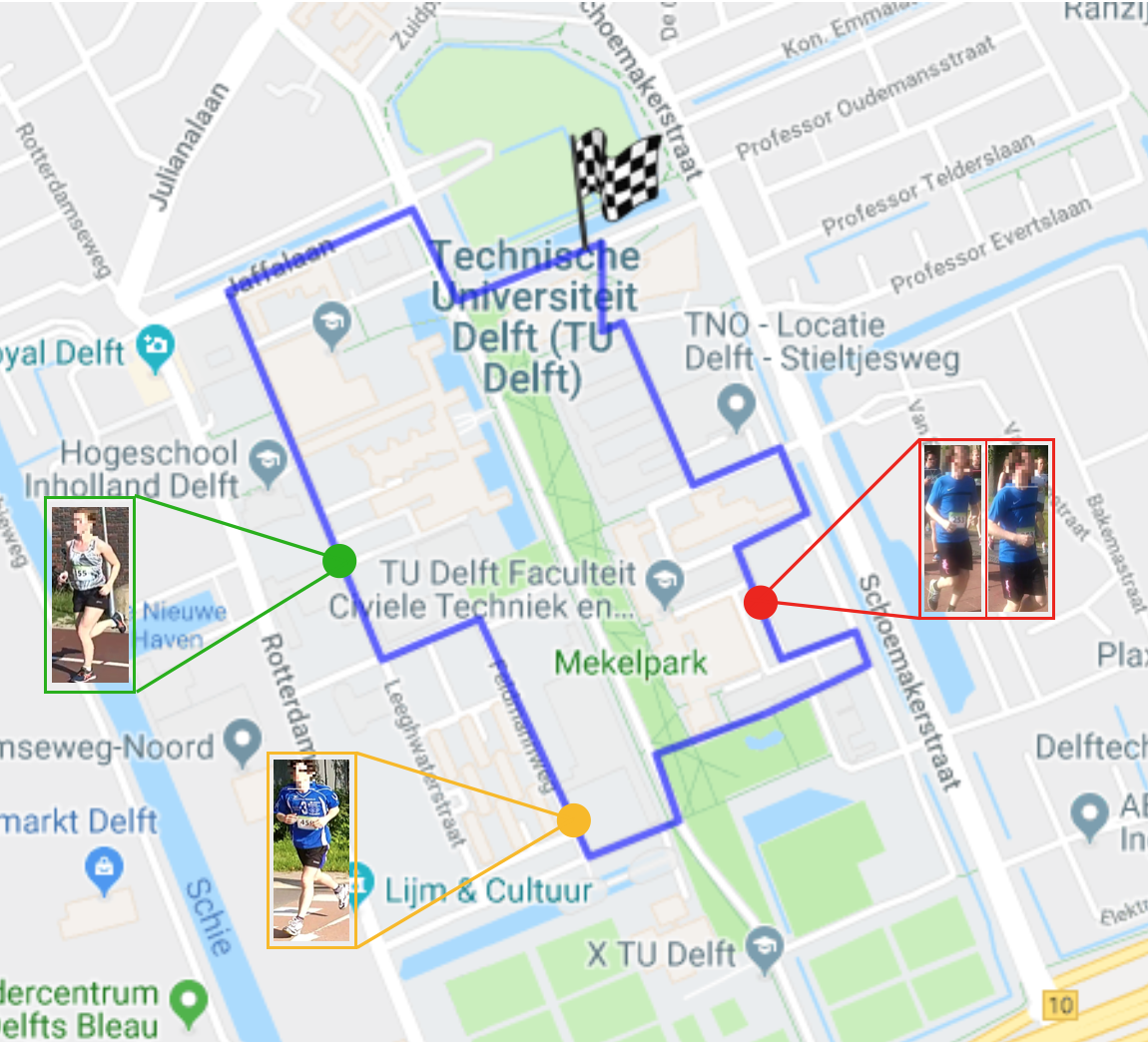}
  \caption{Envisioned result of proposed approach - Shown here is a visualization of a frame of the 2D video with runners represented by markers. The blue line represents the race track where cameras are located at certain points, the flag represents the start and the finish line. The proposed model has the ability to retrieve pictures of runners (at any point of time) given their unique bib number.}
  \label{fig:teaser}
\end{figure}

\begin{figure*}[!ht]
\centering
\begin{subfigure}[t]{.45\textwidth}
  \centering
  \includegraphics[height=4cm,keepaspectratio]{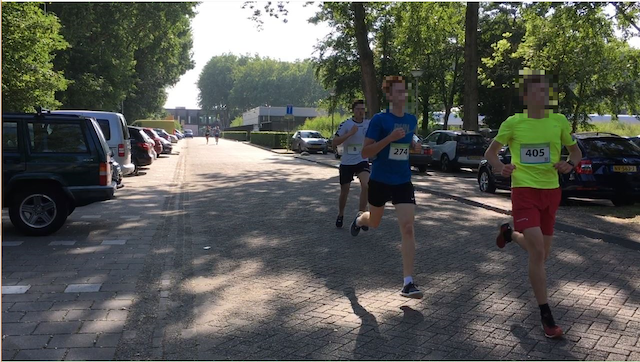}
  \caption{}
  \label{fig:bib_number_good}
\end{subfigure}
\begin{subfigure}[t]{.45\textwidth}
  \centering
  \includegraphics[height=4cm,keepaspectratio]{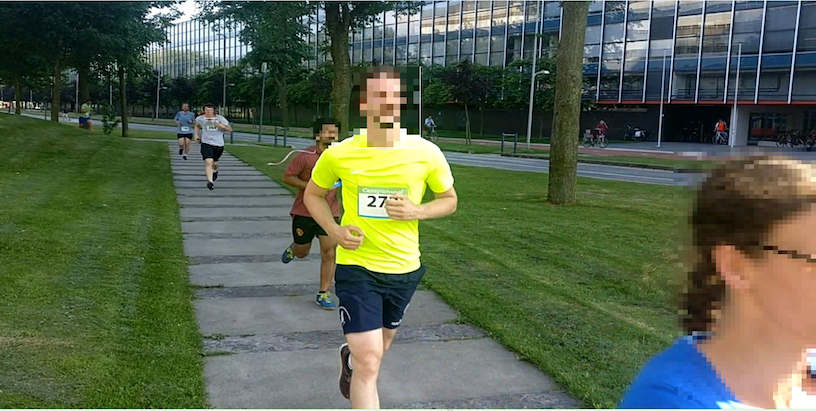}
  \caption{}
  \label{fig:bib_number_occlusion_half}
\end{subfigure}
\caption{Examples of runners with their bib number tag in image samples. (a) Bib number tags with clear visibility: 405 (green T-shirt), 274 (blue T-shirt), and 8 (white T-shirt). (b) Bib number tag 272 has one of its digits occluded; so it looks like a 27. This along with varying lighting conditions, shadows and variance in pose and camera position makes the task of runner detection non-trivial.}
\label{fig:bib_number}
\end{figure*}

\section{Introduction}
Running events have gained more popularity in recent times due to  increasing awareness regarding health and the accessibility of such events for all people regardless of their gender, age, or economic status.

\noindent In some races, the athlete's location data is collected using GPS tracker for live tracking \cite{bib:pielichaty_2017}. This data is used by event organizers to publish statistics or records of the race, or sometimes for the participants themselves to track their performance. Several vendors for GPS trackers provide a service for data analysis and running race visualization from the race data \cite{bib:racemap, bib:tech4race}. However, athletes usually would like to view images/visualizations of themselves retrieved from the event. Also, GPS trackers are personalized to each athlete while cameras can provide an overview for multiple athletes at once. Videos also provide additional information that can be used for athlete health monitoring.  \\
While the video streams are time-stamped and also have GPS information on where they are recorded, there is no prior knowledge as to which runner appears in each of the videos. Given that each runner has a unique 'bib number' identifying them, attached to the shirt and with the assumption that any two random people are likely to be dissimilar in appearance, computer vision models can be used for recognizing them. Scene text recognition models can be used to identify the bib number and ultimately the athlete. However, there are potential issues as the bib numbers or the athletes themselves can be fully or partially occluded, as shown in Figure \ref{fig:bib_number}. 

\noindent Person re-identification is an approach that utilizes a similarity measure (like Euclidean distance), training a model to retrieve instances of the same person across different cameras \cite{bib:survey_personreid_1, bib:deep_personreid_survey_3}. This approach can potentially improve runner identification in scenarios where the bib number is not visible but the runner's features are.  \\
In this work, we use scene text recognition both as a baseline and to collect training samples for a person re-identification method downstream in our pipeline for runner identification. The envisioned result of the model is shown in Figure 1, the race track map with the trajectories of athletes with the ability to retrieve instances of detection of each athlete. The only annotations required are bib numbers that appear in the video to evaluate the scene text recognition model.  For the scene text recognition method, we choose deep-learning based text spotter baselines \cite{bib:deeptextspotter, bib:he_textspotter}.
To evaluate the performance of the proposed approach, we created a ground truth database where the videos are annotated with the bib numbers of appearing runners and the frame interval which they belong.

\noindent Our main contributions are:
\begin{itemize}
    \item A method for visualizing running events using videos recorded at the event. Computer vision based runner detection methods are used with video metadata to find video files and frames where each of the runners appears. The only supervision used in runner identification is the bib numbers known to occur in the video. However, for the whole running event visualization GPS correction and additional filtering steps are used.
    
    \item Evaluation of relevant computer vision models for the detection of runners in videos.
\end{itemize}

\section{Related Work}

\noindent \textbf{Scene Text Recognition} \ Scene text recognition is the detection, localization and identification of text in images. Recognizing the text in images of scenes can be useful for a wide range of computer vision applications, e.g. robot navigation or industry automation \cite{bib:navigation, bib:automation}. Traditionally, handcrafted features (e.g. HOG, color, stroke-width, etc.) are utilized to implement a text recognition system \cite{bib:handcraft_1, bib:handcraft_2}. However, because of limited representation ability of handcrafted features, these methods struggle to handle more complex text recognition challenges \cite{bib:survey_textdetection_1}, like the ICDAR 2015 dataset \cite{bib:icdar_2015}. Deep learning-based methods \cite{bib:deeptextspotter, bib:mask_spotter, bib:he_textspotter, bib:fots} are more successful in this regard because they can learn features automatically. \citet{bib:mask_spotter} modified the Mask R-CNN to have shape-free text recognition ability. \citet{bib:fots} and \citet{bib:deeptextspotter} adopted EAST \cite{bib:east} and YOLOv2 \cite{bib:yolov2} as their detection branch and developed CTC-based text recognition branch. \citet{bib:he_textspotter} adopted EAST \cite{bib:east} as its text detection branch and developed the attention-based recognition branch. Since in our work, the bib number text appears in unconstrained images of scenes (e.g. dynamic background textures, varying size of text due to distance to camera, illumination variation, etc.) as shown in Figure \ref{fig:bib_number}, deep learning based scene text recognition methods would be more well suited to detect the bib numbers because it can perform better with the varying scenarios than traditional (handcrafted features) methods. \\

\noindent \textbf{Bib Number Detection} \ Using scene text recognition methods is an intuitive choice to detect the bib number and consequently identify the runner who wears it. \citet{bib:racing_bibnum} proposed a pipeline consisting of face detection and Stroke Width Transform (SWT) to find the location and region of bib number tag, and then applied digit pre-processing and an Optical Character Recognition (OCR) engine \cite{bib:tesseract} to recognize the bib number text. \citet{bib:svm_bibnumber} proposed to use torso detection and a Histogram of Oriented Gradient (HOG) based text detector to find the location and region of bib number tag, and then they use text binarization and OCR  \cite{bib:tesseract} to recognize the bib number text. Since there is no 'learning' component, the aforementioned models cannot adapt to varying environmental conditions. Existing and publicly available text recognition models \cite{bib:deeptextspotter, bib:he_textspotter} are used as our scene text recognition baseline to detect the bib numbers (so as to not reinvent the wheel - focusing on implementing a text recognition system). \\

\noindent \textbf{Person Re-identification} \ Person re-identification (or re-id) is the task of recognizing an image of a person over a network of video surveillance cameras with possibly non-overlapping fields of view \cite{bib:survey_personreid_1}. In previous research, authors \cite{bib:classification_personreid_2, bib:classification_personreid_1, bib:classification_personreid_3} have investigated classification-based CNN models for person re-id. However, if the dataset lacks training samples per individual, the classification-based model is prone to overfitting \cite{bib:survey_personreid_1}. To counter this, the task is cast as a metric learning problem with Siamese networks \cite{bib:siamese_network}. \citet{bib:pairwise_model_1} uses a Siamese network with pairwise loss that takes a pair of images to train the model and learn the similarity distance between two images. \citet{bib:triplet_loss} showed that using the Siamese network model and triplet loss are a great tool for a person re-identification task because it learns to minimize the distance of a positive pair and maximize the distance of a negative pair at once. \citet{bib:bagoftricks} improved the previous work (Siamese network model with triplet loss) performance with several training tricks, such as center loss or label smoothing. This work \cite{bib:bagoftricks}, is what we adopt as our person re-id model for runner detection.

\section{Methodology}

\subsection{Running Event Visualization}

This section outlines  steps for visualization of the running event. 

\subsubsection{Running track} To visualize computed trajectories of  athletes, a map of the event running track is required. The track is formed by an array of GPS coordinates on an interactive map. The sequence of GPS coordinates are created using \textit{geojson.io}, a  web-based tool to create geo-spatial data in GeoJSON file format \cite{bib:geojson_file}. It provides an interactive editor to create, view, and share maps \cite{bib:geojson_io} similar to Figure \ref{fig:teaser}.

\subsubsection{Runner timestamp}
Another important element to visualize the athletes trajectory is the timestamp where runners appear at different locations. The individual timestamp is determined by the video timestamp and the frame where the runner appears in the video. From video metadata, we use the date last modified $t_{V}$ and Duration $T_{V}$ to get the time the video starts recording. Then we convert the frame $f$ where the runner appears into seconds by dividing it with video frame rate $r$. To get the individual timestamp, we use linear interpolation, adding the video start time and the converted frame. Individual timestamp is given by: 

\begin{equation}
\label{eq:timestamp}
 t_{R} \, = \, t_{V} \, - \, T_{d} \, +  \, \frac{f}{r} \ ,  
\end{equation}

\noindent where $t_{R}$ is the individual athlete's timestamp, $t_{V}$ is the video timestamp, $f$ is the frame where the runner appears, $r$ is the frame rate of the video, and $T_{d}$ is the total duration of the video. 

\begin{figure}[!ht]
\centering
\includegraphics[scale=0.3]{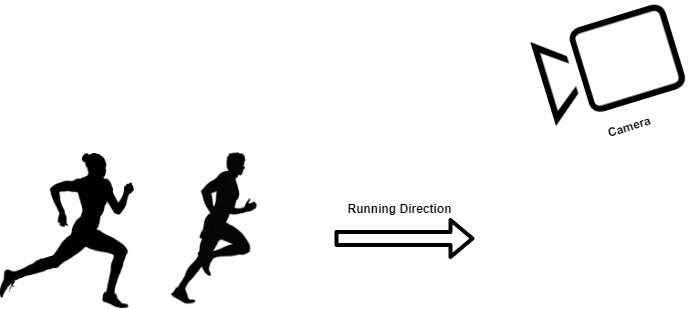}
\caption{Illustration of how video recording was configured . In most videos, the runners move towards the camera. At first, there is considerable distance between the camera and the runners, but the runners progressively get closer to the camera.}
\label{fig:cam_runner_illustration}
\end{figure}

\noindent Since runners can appear in multiple frames, we also take into account where the relative position of runners to the camera to determine the best frame to use. Based on our observation, in most videos, the position of the camera and the athletes resembles Figure \ref{fig:cam_runner_illustration}, in which the runners ran towards the camera. So the last frame where a runner appears would be the best choice (closest to the camera), given that the GPS coordinates of the camera will be used to estimate runner location.

\subsubsection{Filtering raw GPS data}
\begin{figure}[h!]
\centering
\begin{subfigure}{.5\linewidth}
  \centering
  \includegraphics[width=4.5cm,keepaspectratio]{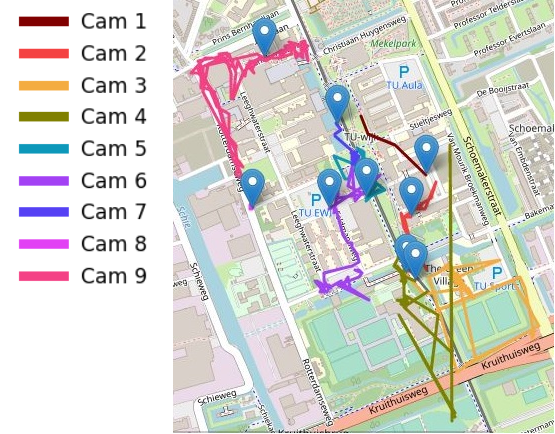}
  \caption{}
  \label{fig:raw_geo}
\end{subfigure}%
\begin{subfigure}{.5\linewidth}
  \centering
  \includegraphics[width=3.1cm,keepaspectratio]{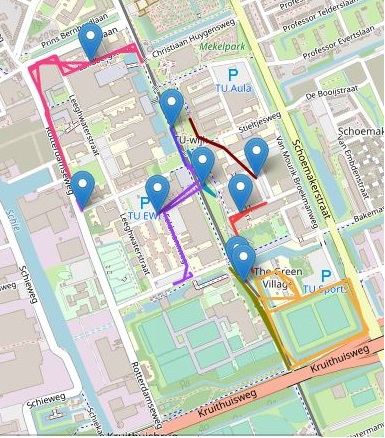}
  \caption{}
  \label{fig:filter_geo}
\end{subfigure}
\caption{Visualization of camera GPS data. (a) Raw camera trajectory. (b) Filtered camera trajectory. There are nine lines with different color representing different cameras. The cameras moved according to the line drawn on the map.}
\label{fig:geotag_data}
\end{figure}

The Figure \ref{fig:raw_geo} shows the trajectory of raw GPS data of videos collected by nine cameras, and it can be seen that the cameras did not stay at one location. Although the videos were taken on the running track, the GPS coordinates seem to stray away. Some of the GPS coordinates deviate too far from the original location; sometimes it is on a highway, a river or a building. The raw GPS data needs to be filtered so that it can be used for visualization.

\noindent The first filtering step of raw GPS data is quite simple, i.e. replacing the raw GPS coordinates with the nearest running track points, so the video GPS coordinates stay in the running track. Cosine law is used to filter stray points based on the angle and distance formed by the stray point and two neighbouring ones.

\noindent After we apply the filters on raw GPS data, we have a better-looking camera trajectory, as shown in Figure \ref{fig:filter_geo}. However, there are still a few stray points from camera 6 that intersects with the trajectory of camera 5 and 7. We manually predict and replace the remaining stray points with new GPS coordinates with estimated points after viewing and analyzing the recorded videos.

\subsubsection{Start and finish location}
A piece of information missing from the video data is the individual timestamp and GPS coordinates at the start and finish because there is no video recorded at both locations. So the GPS coordinates and the individual timestamp at the start and finish location are determined by ourselves. The GPS coordinates at the start and finish location are predicted based on the start and finish locations provided by the event organizer. Meanwhile, the individual timestamp of the start location is determined by setting to zero the seconds of the individual timestamp at the location of the first camera. For example, if the individual timestamp at camera 1 is 16:00:10, then the start timestamp is 16:00:00. We assumed that the camera 1 location is only a few meters away (10 - 20 meters) from the start location. The individual timestamp at the finish location is defined by the start time plus the duration required by a runner to finish the race. This duration is obtained from scraping the Campus Run result website. After the timestamps and the GPS coordinates of the start, finish, and camera locations are obtained, the runner's motion can be interpolated between those locations. Then the javascript library, D3\footnote{https://d3js.org/} and Leaflet\footnote{https://leafletjs.com/} are used to create the visualization.

\subsection{Runner Detection}

The model needs to retrieve the video and frame information where the athletes are detected. We are interested in that information because they are used to visualize the runner's trajectory; the video provides GPS coordinates and video timestamp information, and the frame represents the individual timestamp.

\subsubsection{Scene text recognition}
The steps to use the text spotter model \cite{bib:deeptextspotter, bib:he_textspotter} for our runner detection task are :

\begin{itemize}
    \item the frames of a video are fed into text spotter model,
    \item all texts in the image are detected by the text spotter, 
    \item if the bib number we know from scrapping the marathon website is detected, we retrieve the video and frame information for further evaluation.
\end{itemize}

\noindent Text spotter also collects training samples for person re-id which are cropped person images from the video dataset. The cropping occurs around a specified region if the text spotter detects a bib number on the image. Then the bib number is assigned as the label of that cropped image. 

\noindent Although there are two text spotters, we do not merge the training samples collected by the two text recognition models. Instead, the person re-id model is trained with each training sample separately so as to compare the results of both models.

\subsubsection{Person Re-identification}
Given the effectiveness of metric learning \cite{bib:triplet_loss}, we choose to adopt the work of \citet{bib:bagoftricks} which proposes a Siamese network with triplet loss and cross entropy loss.

\noindent Since we do not have the ground truth identities of athletes annotated, to evaluate the person re-id model, an object detector is used to detect people in the video dataset and these detections are cropped out. Those cropped images do not have a label because the object detector cannot detect the bib numbers, but the video and information are still stored. The person re-id model is evaluated to see if it can recognize the people detected by the object detector.

\noindent It is possible that the object detector results are noisy and might include pedestrians in addition to athletes. So, a \textit{k-NN} is used as an outlier detector, it has competitive performance compared to other outlier detection methods \cite{bib:survey_outlier_detection}.

\noindent The idea of using \textit{k-NN} as an outlier detector is that the larger the distance of a query point from its neighbor points, higher the likelihood that it is an outlier \cite{bib:knn_distance}. We adopt the definition of an outlier that considers the average of the distance from its $k$ neighbors \cite{bib:knn_outlier}. If the average distance of an image from its $k$ neighbors is larger than a threshold, then it is considered as an outlier.

\subsection{Performance Evaluation}
This section describes the mathematical formulation of the evaluation metrics and their implementation in our problems.

\subsubsection{Video-wise metric}
The video-wise metric is useful to check the number of videos where a runner is detected at least once. Evaluating the performance on the retrieved video information is important because we use the video GPS coordinates to visualize the runners trajectory, so we want to retrieve as many relevant videos as possible. At the same time, it is also undesirable to retrieve irrelevant videos because the irrelevant videos might have GPS coordinates and timestamps that do not agree with GPS coordinates and timestamps that the relevant videos have. Therefore, it could hinder the runner's trajectory visualization.

\noindent True positives are when a positive class is correctly predicted to be so, and false positives are when a negative class is detected as positive \cite{bib:olson_delen_2008}. In our problem, we collect a ground truth database consisting of the video filename, bib number, and frame interval. So true positive could be defined as correctly retrieved information (video filename, bib number, frame), and the false positive for incorrectly retrieved ones. To evaluate the video-wise metric, the retrieved information we need is only the bib number and its video. We use F1-score for the video-wise metric to evaluate on both relevant and irrelevant videos retrieved by runner detection methods. The video-wise F1-score formula is defined as followed:

\begin{equation}
\label{eq:recall}
\text{Recall}_{i}^{v} \, = \frac{|\text{TP}|_{i}^{v}}{|\text{GT}|_{i}^{v}} \ ,  
\end{equation}

\begin{equation}
\label{eq:precision}
\text{Precision}_{i}^{v} \, = \frac{|\text{TP}|_{i}^{v}}{|\text{TP}|_{i}^{v} + |\text{FP}|_{i}^{v}} \ ,  
\end{equation}

\begin{equation}
\label{eq:f1_score}
\text{F1-score}_{i}^{v} \, = 2 \ . \ \frac{\text{Recall}_{i}^{v} \ . \ \text{Precision}_{i}^{v}}{\text{Recall}_{i}^{v} + \text{Precision}_{i}^{v}} \, , \end{equation}

\noindent where $i$ denotes the runner, $v$ denote a video-wise metric, $|GT|$ denotes the number of runner's video in ground truth, $|TP|$ denotes the total true positive (relevant) videos and $|FP|$ denotes the total false positive (irrelevant) video. The final score is defined as the average over $M$ classes, as shown below :

\begin{equation}
\label{eq:average_f1}
\text{F1-score}^{v} \, = \frac{1}{M} \sum_{i=1}^{M} \, \text{F1-score}_{i}^{v}  \ .
\end{equation}

\subsubsection{Frame-wise metric}
Since it is possible that every one of the runners are seen at least once during the event, a naive method that predicts every athlete to be visible in every camera at all times, could achieve a high F1-score. So we need a frame-wise metric. Also, runner detection methods might produce false positives at an irrelevant frame in a relevant video. So, it is necessary to evaluate at a frame level in addition to the video level evaluation. 

\noindent We use temporal Intersection of Union (IoU) for the frame-wise metric, which measures the relevant interval of frames where the runner appears. The formula of temporal IoU is similar to regional IoU used in object detection \cite{bib:iou_jaccard}, except it is only one-dimensional, as shown in Figure \ref{fig:temporal_iou_1d}. The formula for calculating temporal IoU is defined as followed:

\begin{figure}[]
\centering
\includegraphics[width=1.\linewidth, keepaspectratio]{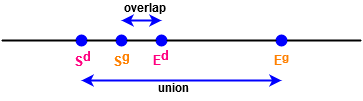}
\caption{Visualization of frame interval intersection. $S$ denotes the frame start, and $E$ denotes the frame end. And $d$ denotes that the frame interval belongs to detection result and $g$ denotes that the frame interval belongs to the ground truth. The temporal IoU is the ratio between the width of overlap and union.}
\label{fig:temporal_iou_1d}
\end{figure}

\begin{equation}
\label{eq:overlap}
\text{overlap}_{ij} =  
    \min(E_{ij}^{g} \, , \, E_{ij}^{d}) - \max(S_{ij}^{g} \, , \, S_{ij}^{d}) \, ,
\end{equation}

\begin{equation}
\label{eq:union}
\text{union}_{ij} \, = (E_{ij}^{g} - S_{ij}^{g}) \, + \, (E_{ij}^{d} - S_{ij}^{d}) - \text{overlap}_{ij} \ ,
\end{equation}

\begin{equation}
\label{eq:iou_class}
\text{IoU}_{ij} = 
    \begin{cases}
        0 \, ,& \text{if overlap}_{ij} < 0 \\ 
        \frac{\text{overlap}_{ij}}{\text{union}_{ij}} \, ,& \text{otherwise}
    \end{cases}
\end{equation}

\begin{equation}
\label{eq:iou}
    \text{IoU}_{i} = \frac{1}{N_i} \sum_{j=1}^{N_i} \, \text{IoU}_{ij}  \ ,
\end{equation}

\noindent where $i$ denotes the runner, $j$ denotes the video file, and $N_{i}$ is the total of video files where the runner detected. $S_{ij}^{g}$ and $E_{ij}^{g}$ are the frame start and the frame end of runner $i$ at video $j$ from ground truth database. $S_{ij}^{d}$ and $E_{ij}^{d}$ is the first frame and the last frame which the runner $i$ is detected at video $j$. The temporal IoU per runner is the average of the temporal IoU over the total of retrieved video. If the detection happens at an irrelevant video $j$, then $IoU_{ij}$ is zero, so it will lower the $IoU_i$. The final score is calculated as average over $M$ classes, expressed as :

\begin{equation}
\label{eq:average_iou}
    \text{mIoU} = \frac{1}{M} \sum_{i=1}^{M} \, \text{IoU}_{i}  \ .
\end{equation}

\begin{figure*}[htp!]
\centering
\includegraphics[scale=0.4]{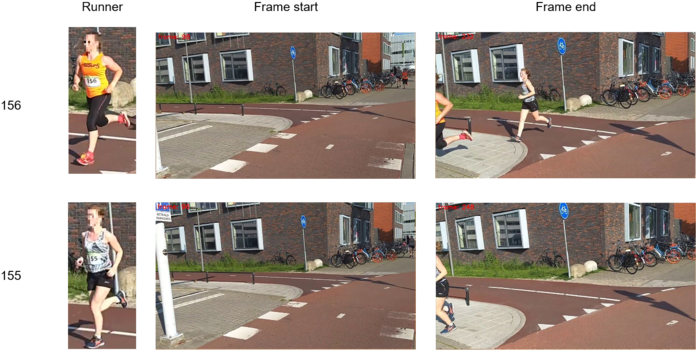}
\caption{An example of video annotation. The video is annotated with the runners appearing within a frame interval. The frame interval consists of frame start where the runner start appearing and frame end where the runner is out of frame.}
\label{fig:labeling_2}
\end{figure*}

\section{Experiments}

\subsection{Datasets}
\subsubsection{Ground truth database}
\label{sec:ground_truth}

Ground truth is necessary for evaluating the performance of the model. It is obtained by manual video analysis, defining a range of frames for each runner where they are recognizable by the human eye. Due to the massive amount of data and time constraints, only runners from the 5 km category are fully annotated. Consequently, only videos with 5 km runners (a total of 127 people) are annotated as ground truth for evaluation. There are a total of 127 runners for the 5km category. 187 videos of these runners were recorded across 9 different locations. The average length of these videos are 35.44 seconds.  

\noindent The ground truth is established by a range of frames where the runners appear. The range is defined by the "frame start" and "frame end" for every runner in the video. The frame start is annotated when a runner is first recognizable by human-eye. Meanwhile, the frame end is annotated when a runner exits the camera field of view. Figure \ref{fig:labeling_2} shows that runner 156 is first seen at the right edge of the screen at frame 98, so this frame is annotated as frame start for runner 156. Then at frame 232, runner 156 is seen for last time at the video, then this frame is annotated as frame end for runner 156. Meanwhile, for runner 155, the frame start is 96, and the frame end is 248. 

\subsubsection{Evaluation set for person re-id} 

The person re-id model can not localize and recognize a person simultaneously from a given frame containing multiple objects and the background. It requires cropped images with one person in each image. The training and evaluation set thus must be a collection of cropped images of people, not videos. To create this evaluation set for the person re-id model , images of people from the videos are collected by an object detector and cropped. The difference between the person images collected by the object detector and that collected by text spotters is that the images from the object detector do not have a label. So, the object detector identifies any images with people from a video regardless of whether the person has a bib number or runners with occluded bib number text. The video and frame information of the extracted images of people are also stored for performance evaluation.

\noindent There are many object detection methods, such as Faster R-CNN \cite{bib:faster_rcnn}  R-FCN \cite{bib:rfcn} and single shot detector \cite{bib:ssd}. However, experiments comparing them \cite{bib:survey_object_detectors} show that Faster R-CNN has better accuracy compared to the other two, so Faster R-CNN is chosen.

\subsection{Implementation Details}

\noindent For text spotter \cite{bib:deeptextspotter, bib:he_textspotter} and the object detector Faster R-CNN, available pre-trained models are used. For the person re-id model, we use the same configuration options as the author \cite{bib:bagoftricks}, using the Adam optimizer \cite{bib:gradient_descent} and the same number of total epochs (120). The output from the object detector (cropped images of people) with the output of scene text recognition as annotation (bib numbers identify runners) is fed into the person re-id model.

\noindent Additionally, we collect new training samples to retrain the person re-id model from the images of people obtained using the object detector. We choose the cropped images that are considered as inliers by the previous round of training with the person re-id model.

\noindent For \textit{k-NN} model, we choose $k$ = 5 and use cosine distance as its distance metric during the inference stage as it is shown to be better than using Euclidean distance for the person re-id task. \cite{bib:bagoftricks}.

\noindent The \textit{k-NN} model is trained on the embedding features extracted by the person re-id model. The person re-id model is used only as a feature extractor.

\subsection{Performance Analysis}

\subsubsection{Bib number detection}

\begin{figure}[!ht]
\centering
\includegraphics[width=1\linewidth,keepaspectratio]{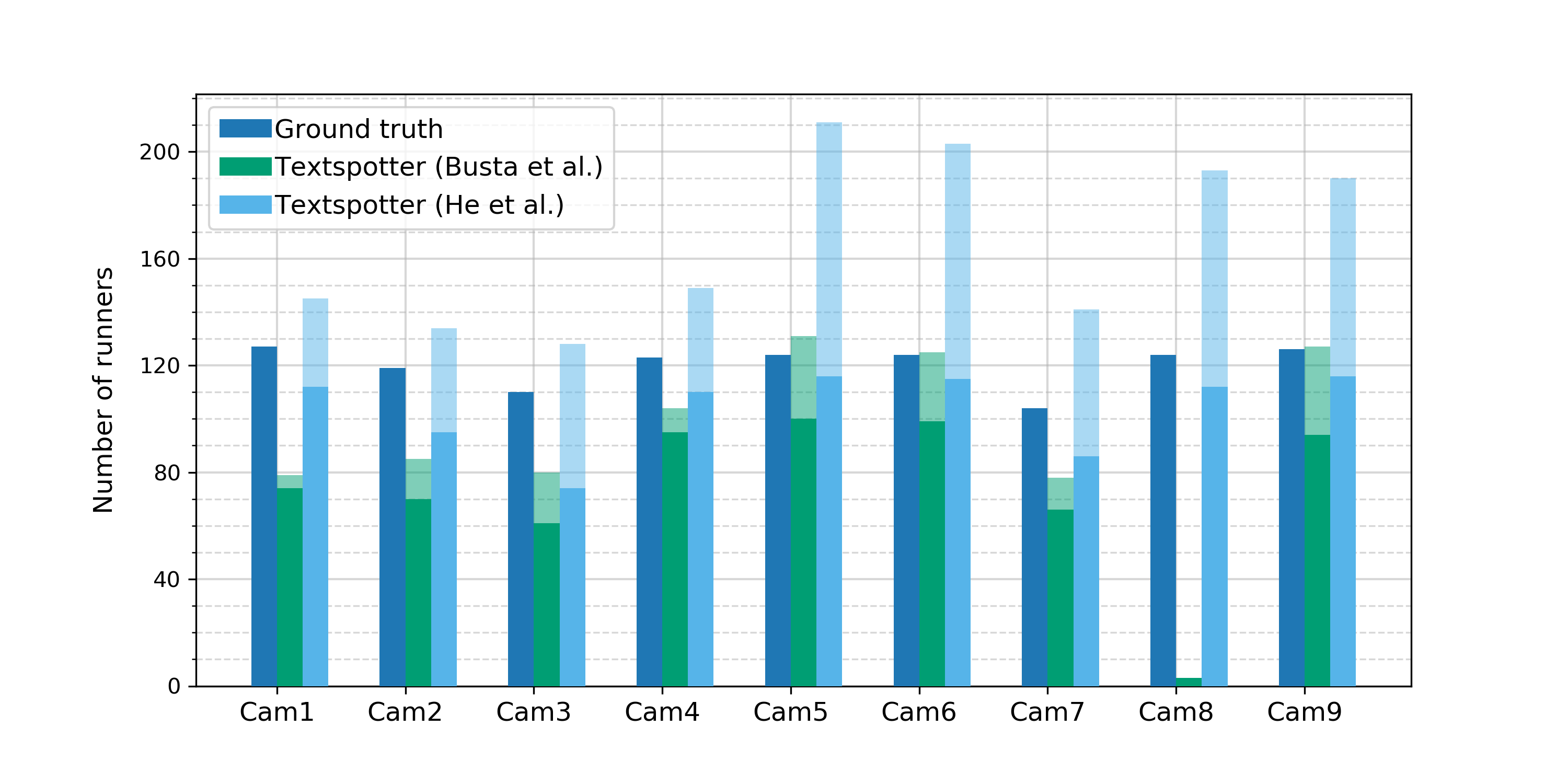}
\caption{The comparison of the number of runners detected per camera on the event track. The stacked bars with lighter colors are the false positives produced by the text spotter. \citet{bib:he_textspotter} has a higher number of true positives and false positives than \citet{bib:deeptextspotter}.}
\label{fig:num_runner_textspotter}
\end{figure}

The model proposed by \citet{bib:he_textspotter} is better at detecting the bib numbers than \citet{bib:deeptextspotter}. \citet{bib:he_textspotter} has larger recall but lower precision compared to \citet{bib:deeptextspotter}, as shown in Table \ref{tab:textspotter_recall_precision}. Consequently, although \citet{bib:he_textspotter} detects more runners, it produces a larger number of false positives than \citet{bib:deeptextspotter}, as shown in Figure \ref{fig:num_runner_textspotter}. Even if a few digits of bib number is occluded, the text spotter \cite{bib:he_textspotter} will detect it. But because the occluded bib number could look like another number, it will count as false positive. These methods are also compared against naive baselines (Table (1)), that predict all runners to be visible at all times in each video (Baseline (all)) and a random prediction given the number of visible people in the frame (Baseline (random)). Per video, the Baseline(all) has a high recall owing to the fact that all runners are predicted to be visible at all times, this resulting in zero false negatives. However, there are a lot of false positives as not all runners are visible in every video at all times.

\noindent The number of detected runners by text spotter \cite{bib:deeptextspotter} from camera 8 is quite low probably because videos recorded by 8 are recorded under harsher illumination, which could interfere with the text spotters \cite{bib:deeptextspotter} performance.

\begin{table}[!htp]
\caption{The average performance of text spotter on our dataset. \citet{bib:he_textspotter} has higher recall, but lower precision compared to \citet{bib:deeptextspotter}.}
\begin{tabular}{@{}cccc@{}}
\toprule
\textbf{Text Spotter} & \textbf{Recall$^{v}$} & \textbf{Precision$^{v}$} & \textbf{F1-score$^{v}$} \\ \midrule
\citet{bib:deeptextspotter} & 61.03 & 76.85 & 66.64 \\
\citet{bib:he_textspotter} & 86.41 & 68.41 & 74.05 \\ 
Baseline (all) & 100 & 40.52 & 57.66 \\
Baseline (random) & 12.76 & 14.49 & 13.57 \\\bottomrule
\end{tabular}
\label{tab:textspotter_recall_precision}
\end{table}

\subsubsection{Distance threshold for person re-id}
A threshold is determined to separate the inliers and outliers for the person re-id methods. We use F1-score$^v$ as a comparison metric between different thresholds. Then we choose a threshold that gives the highest F1-score$^v$. For person re-id with training samples from \citet{bib:deeptextspotter}, we choose a threshold of 0.21 that gives an average F1-score$^v$ of 79.00. Also, for person re-id with training samples from \citet{bib:he_textspotter}, we choose a threshold of 0.22 that gives an average F1-score$^v$ of 85.77. 

\noindent Since the person re-id with training samples collected by \citet{bib:he_textspotter} has higher performance, the detection results from this person re-id model are used to train the model for the second time. For this retrained person re-id model, with 0.05 as the distance threshold an average F1-score$^v$ of 87.76 is obtained. 

\subsubsection{Comparative results: F1-score$^v$}
\begin{figure}[!htp]
\centering
\includegraphics[width=\linewidth, keepaspectratio]{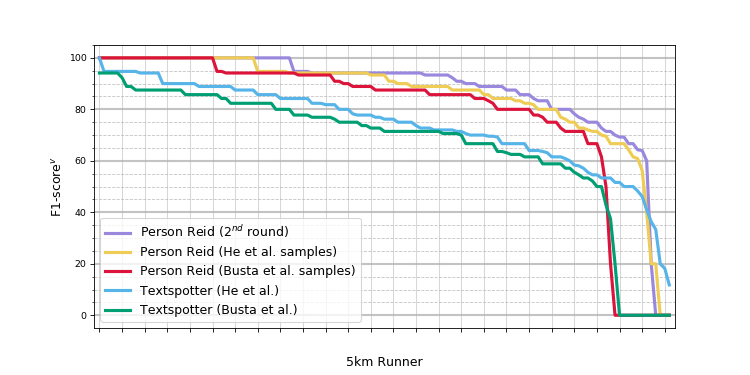}
\caption{The comparison of F1-scores$^{v_{i}}$ between person re-identification and text spotter models per runner. The x-axis presents runners sorted on F1-score$^{v_{i}}$. The x-axis values are not shown because there are five different axes. Based on F1-score$^{v_{i}}$,  person re-id models outperform the text spotters.}
\label{fig:video_f1_plot}
\end{figure}

We also report the F1-score$^v$ between two text recognition models and also the person re-id model trained in three different scenarios. Figure \ref{fig:video_f1_plot} shows that person re-id models generally achieves higher performance compared to the scene text recognition models. It validates the hypothesis that using the whole appearance of a person is better for runner detection as compared to using just the bib number. This could be because of occlusion or blurring of bib numbers (due to athlete motion) hindering the performance of text recognition models. Meanwhile, based on visual inspection of detection results, partial occlusion on runner's body or bib number tag, different runner's pose, or different camera setting (e.g. camera viewpoint or illumination) does not have any significant effect on the performance of the person re-id model. As long as the appearance of runners is distinguishable, the person re-id model can detect them. Person re-id fails when the recording setup is not ideal (e.g. the distance between the camera and runners is too far, the runner facing against the camera), or a considerable part of the runner's body is occluded.

\noindent Figure \ref{fig:video_f1_plot} also shows a comparison between person re-id models with different training samples. Person re-id with training samples from \citet{bib:he_textspotter} has higher performance compared to another with training samples from \citet{bib:deeptextspotter} because \citet{bib:he_textspotter} has higher recall so it collects more true positive runners for training. Meanwhile, retrained person re-id only improves a little because most remaining undetected runners are the challenging ones (e.g. they are recorded farther away from the camera, and not visually distinguishable from one another). 

\noindent Another important observation is that runners with zero F1-score$^v$ at the lower right side of the plot. They are the runners whose bib numbers have less than three digits, which the text spotter  struggles to detect. Consequently, the person re-id model does not have training samples and thus it also has a performance of zero for those particular runners. 

\subsubsection{Comparative results: temporal IoU}
\begin{figure}[!htp]
\centering
\includegraphics[width=\linewidth, keepaspectratio]{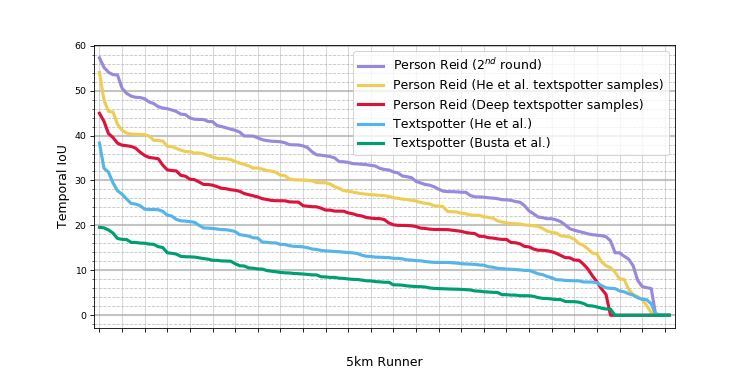}
\caption{The comparison of temporal IoU between person re-identification and text spotter \cite{bib:he_textspotter} per runner. The x-axis is runners sorted on its temporal IoU. The x-axis values are not shown because there are five different x-axes. Person re-id models also exceed the performance of text spotters in terms of temporal IoU.}
\label{fig:temporal_iou_plot}
\end{figure}

To show the performance in retrieving the relevant frames, we present the temporal IoU plot for scene text recognition and three person re-id methods, as shown in Figure \ref{fig:temporal_iou_plot}. The person re-id model still outperforms the scene text recognition ones. It happens because the person re-id uses the training samples from the text spotter, and it can expand the frame interval by detecting runner from earlier frames where the text spotter might find difficulties in detecting the bib number; in the earlier frames, the bib number images are much smaller, but the appearance of the person can be distinct. It is important to notice that some runners have zero temporal IoU. They are the same runners that have zero F1-score$^v$. 

\noindent The analysis of the comparison between person re-id models with different training samples is more or less similar to the analysis on F1-score$^v$, except it is analyzed on a frame by frame basis. \citet{bib:he_textspotter} is better at detecting bib number with a smaller size in the earlier frames; thus person re-id with training samples from this text spotter has a better chance at expanding the frame interval. Meanwhile, the retrained person re-id only improves the average temporal IoU only by 5\%. It happens because the runner's images in remaining frames are the hard ones to detect (e.g. the small image at earlier frames due to the distance) and some false positives remain in the second training set.

\begin{figure}[!h]%
\centering
\includegraphics[width=\linewidth, keepaspectratio]{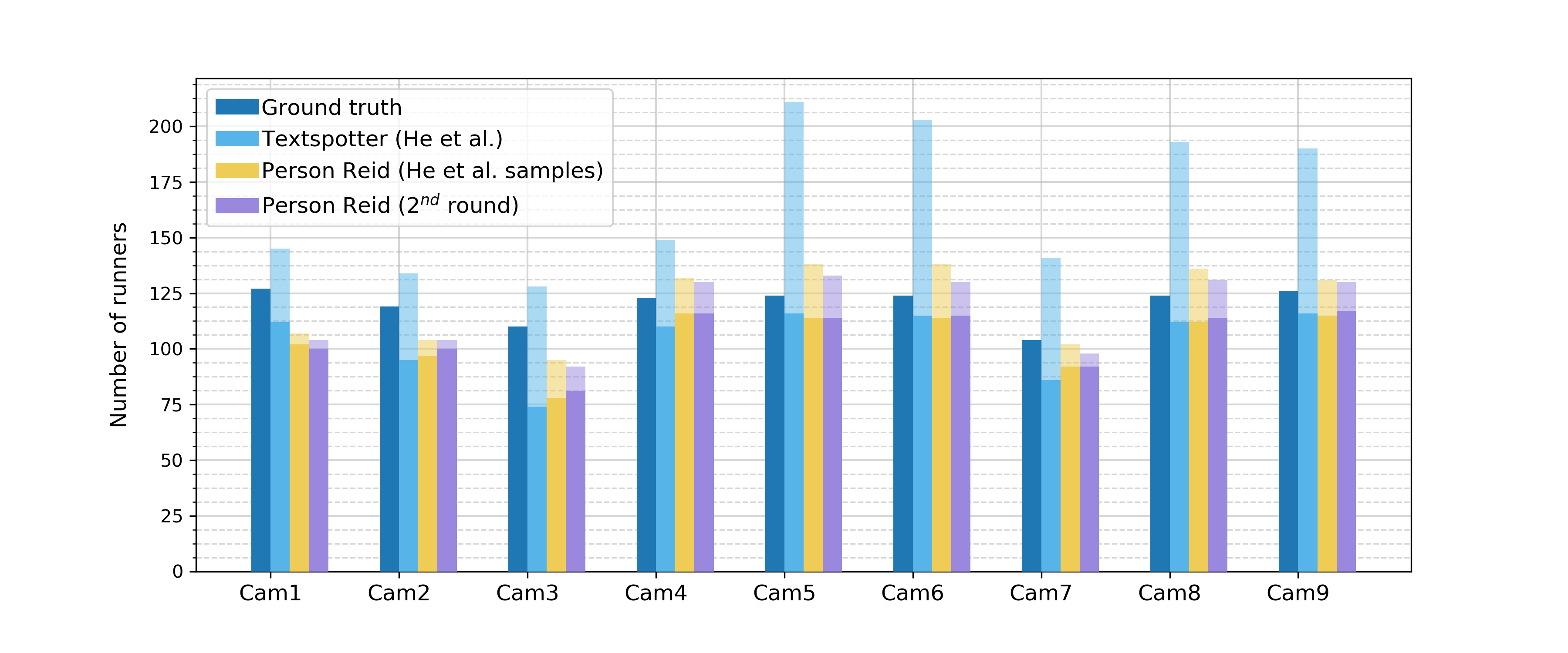}
\caption{The comparison of the number of detected runners per camera between text spotter \cite{bib:he_textspotter} and the corresponding person re-id models. The stacked bars with lighter color are false positives. Person re-id models produce less false positives compared to the text spotter \cite{bib:he_textspotter}.}
\label{fig:num_runner_reid_he}
\end{figure}

\begin{figure*}[!htp]
\centering
\makebox[\textwidth][c]{\includegraphics[scale=0.4]{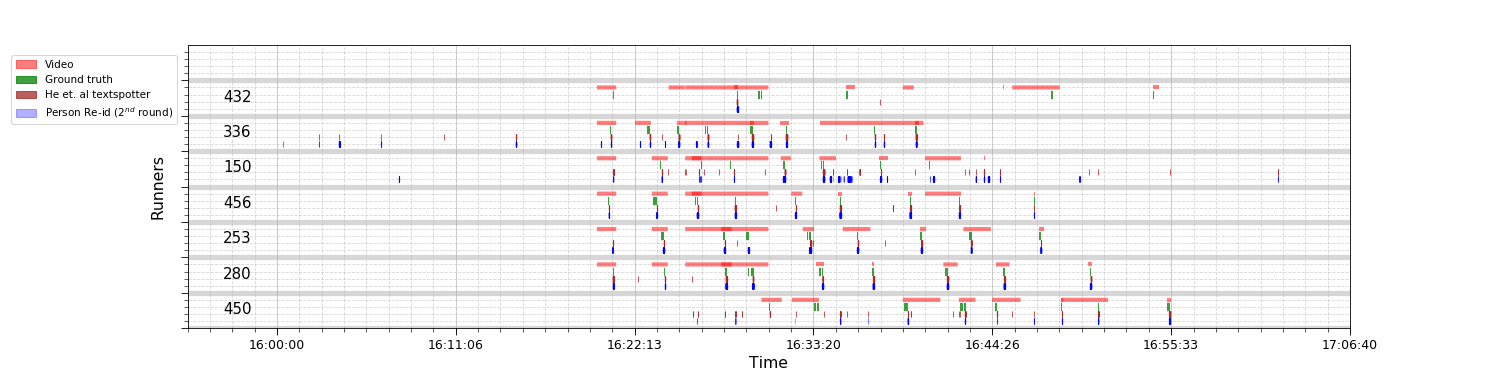}}
\caption{2D timeline visualization of runner detection. The detection result of a person re-id model retrained for the second time and text spotter \cite{bib:he_textspotter} are visualized. Not all runners are visualized, three runners most detected are shown (280, 253, 456) and four runners with the lowest (non-zero) detections (432, 450, 336, 150). True positives are the blue strips (person re-id) or brown strips (scene text recognition) that are aligned with green strips (ground truth) and false positives are defined otherwise.}
\label{fig:timeline_paper}
\end{figure*}
%

\subsubsection{Outlier detection}
In Figure \ref{fig:num_runner_reid_he}, we show the comparison of number of detected runners per camera between \citet{bib:he_textspotter} and the corresponding person re-id models. It can be seen that the number of false positives produced by \citet{bib:he_textspotter} is quite high. However, the person re-id method can significantly improve producing lesser false positives, although it uses the training samples from the text spotter that has many false positives. It validates that \textit{k-NN} in person re-id performs well enough as an outlier detector. Person re-id model with triplet loss minimize the similarity distance between images of the same person and maximize the similarity distance between images of different persons, so the false positives of a runner will have further similarity distance from the training samples of a runner. Then the false positives with larger distances could be rejected.

\noindent However, in the second round of training the person re-id model, the reduction in false positives is not significant. This could be because some of sampling more false positives, some of which may be similar in appearance to the runners, and thus are also not rejected as an outlier.
\subsubsection{2D timeline visualization}
\noindent Figure \ref{fig:timeline_paper} shows the true positives and false positives in our research problem clearly. It can be seen that runners 280, 253, and 456 have all their blue strips aligned with their green strips; using retrained person re-id, they have perfect F1-score$^v$ of 100. Meanwhile, using retrained person re-id, runners 336, 150, and 450 still have many false positives. It occurs because of the false positives from text spotter \cite{bib:he_textspotter} that are used as training samples; the false positives from blue strips are aligned with the false positive from brown strips.

\noindent Another interesting observation is that sometimes the person re-id can detect a runner, although it does not have the training samples at that video. For example, the fourth blue strips at runner 253 do not have aligned brown strips.

\section{Limitations}

It is important to note that the performance of the person re-id method depends heavily on the performance of scene text recognition as the latter collects the training samples for the former. For example, person re-id cannot detect the runners that have the bib number with less than three digits. Another thing to note is that there are false positives in the images retrieved by the text spotter. This hinders the performance of person re-id model. 

\section{Conclusions and Future Work}

In this study, we have proposed an automatic approach to create a running event visualization from the video data. We use scene text recognition and person re-identification models to detect the runners and retrieve the videos and frames information where the runners appear so that we can use the individual timestamp and filtered GPS coordinates from the retrieved data for visualization. The experiments show that the scene text recognition models encounter many challenges in runner detection task, which can be mitigated by a person re-id model. The results also show that the performance of the person re-id method outperforms the scene text recognition method. The person re-id method can retrieve the relevant video information almost as good as the ground truth. 

\noindent This research focused on creating 2D visualizations of athletes with timeline charts and running track visualizations with runners represented by moving markers. This can be further extended to 3D using human pose estimation and spatio-temporal reconstruction \cite{mustafa2016temporally}. Gait information could also be used additionally for identification. Such reconstruction is not only useful for sports performance analysis and health monitoring, but can also be used for forensic investigations \cite{chen2016videos}. It would also be interesting to investigate if using another means to collect training samples for person re-id, such as crowdsourcing labeling, can produce a better performance.

\section{Acknowledgement}
This study is supported by NWO grant P16-28 project 8 Monitor and prevent thermal injuries in endurance and Paralympic sports of the Perspectief Citius Altius Sanius - Injury-free exercise for everyone program.

\newpage

\bibliographystyle{ACM-Reference-Format}
\bibliography{sample-base}

\end{document}